\documentclass[11pt]{article}

\usepackage{acl}

\usepackage{times}
\usepackage{latexsym}

\usepackage[T1]{fontenc}

\usepackage[utf8]{inputenc}
\usepackage{caption}
\usepackage{booktabs}
\usepackage{seqsplit}
\usepackage{booktabs} 
\usepackage{amsmath}
\usepackage{url}
\usepackage{algorithm}
\usepackage{algpseudocode}
\usepackage{booktabs} 
\usepackage{graphicx}
\usepackage{mathtools} 
\usepackage{caption}
\usepackage{float}
\usepackage{siunitx}
\usepackage{amssymb}
\usepackage{microtype}

\usepackage{inconsolata}

\usepackage{graphicx}

\title{M$^{4}$olGen: Multi-Agent, Multi-Stage Molecular Generation under Precise Multi-Property Constraints}

\author{Yizhan Li$^{1,2}$, Florence Cloutier$^{1,2}$, Sifan Wu$^{1,2}$, Ali Parviz$^{2}$, Boris Knyazev$^{1,2,4}$, \\
\textbf{Yan Zhang$^{2,4}$, Glen Berseth$^{1,2,3,5*}$ \& Bang Liu$^{1,2,3,5}$}\thanks{Corresponding authors: Bang Liu \texttt{<bang.liu@umontreal.ca>} and Glen Berseth \texttt{<glen.berseth@umontreal.ca>}.} \\
$^1$DIRO \& Université de Montréal \\
$^2$Mila – Quebec AI Institute \\
$^3$Institut Courtois \\
$^4$Samsung AI Lab, Montreal \\
$^5$Canada CIFAR AI Chair \\
}

\begin{document}
\maketitle
\begin{abstract}
Generating molecules that satisfy precise numeric constraints over multiple physicochemical properties is critical and challenging. Although large language models (LLMs) are expressive, they struggle with precise multi-objective control and numeric reasoning without external structure and feedback. We introduce \textbf{M$^{4}$olGen}, a fragment-level, retrieval-augmented, two-stage framework for molecule generation under multi-property constraints. \textbf{Stage~I: Prototype generation}: a multi-agent reasoner performs retrieval-anchored, fragment-level edits to produce a candidate near the feasible region. \textbf{Stage~II: RL-based fine-grained optimization}: a fragment-level optimizer trained with Group Relative Policy Optimization (GRPO) applies one- or multi-hop refinements to explicitly minimize the property errors toward our target while regulating edit complexity and deviation from the prototype. A large, automatically curated dataset with reasoning chains of fragment edits and measured property deltas underpins both stages, enabling deterministic, reproducible supervision and controllable multi-hop reasoning. Unlike prior work, our framework better reasons about molecules by leveraging fragments and supports controllable refinement toward numeric targets. Experiments on generation under two sets of property constraints (QED, LogP, Molecular Weight and HOMO, LUMO) show consistent gains in validity and precise satisfaction of multi-property targets, outperforming strong LLMs and graph-based algorithms. 
\end{abstract}

\section{Introduction}
Generating molecules that satisfy precise numeric constraints is a fundamental and critical task in scientific discovery, with applications in drug development, materials design, de novo design and molecular property optimization~\citep{doi:10.1126/science.aat2663,Fromer_2023}. Optimizing compounds to meet numeric multi-property targets improves real development outcomes with desired attributes~\citep{wager2016central,m2024augmenting}. Much of the molecular generation literature treats molecular discovery as maximizing one or a few surrogate properties, rather than matching user-specified numerical targets; approaches that offer precise, simultaneous control over multiple properties remain scarce.

While numerous application-specific target properties can be used in practice~\citep{MatterGen2025,ding2024matexpertdecomposingmaterialsdiscovery}, for our study we pick fundamental properties often used in prior work~\citep{loeffler2024reinvent,brown2019guacamol,jain2023multi,cai2024foundation}.
Specifically, we begin our study with simple properties, namely drug-likeness (QED), lipophilicity (logP), and molecular weight (MW) that shape permeability, exposure, and overall developability~\citep{bickerton2012quantifying,giaginis2018impact}. 
While these are simplified surrogates, they are (i) fast and reproducible to evaluate (enabling large-scale training and ablations), (ii) continuous and numeric, which is essential for testing precise multi-objective control, and (iii) standardized across open benchmarks, supporting fair comparison. 
We then conduct more challenging experiments on energy properties, HOMO and LUMO~\citep{bredas2017organic, fukui1954molecular}, critical in diverse applications~\citep{kim2013energy}. Our goal is to validate a multi-agent, numerically conditioned generation framework under multiple verifiable, compute-efficient proxies. As we show with HOMO-LUMO experiments, our approach can be used in richer oracles as we scale to more realistic discovery settings. 

To fine-tune the optimizer, we construct a large dataset of more than 2 million molecules decomposed into BRICS~\citep{degen2008art} fragments along with their corresponding properties. From this dataset, we derive a neighbor relational dataset of 1.17 million pairs for controllable reasoning automatically. Each molecule in this dataset is paired with an explicit one-hop neighbor list: molecules that differ by exactly one fragment (add, remove, or replace) and that pass the RDKit validity and edit sanity checks. By chaining these one-hop moves, we gradually grow neighbor forests from any starting molecule. These structures enable long and controllable reasoning chains.
We demonstrate that this architecture markedly improves adherence to numeric multi-property constraints and surpasses prior LLM-based methods by large margins. 

In summary, we contribute (i) M$^{4}$olGen, a molecular generation framework that couples retrieval-augmented prototyping with GRPO-based fragment-level optimization to achieve \emph{exact numeric control} over multiple properties; (ii) a scalable \emph{multi-hop} refinement mechanism that boosts output quality while explicitly regulating edit complexity and deviation from the starting structure; (iii) a public dataset of $\sim$2.95M molecules with BRICS fragment annotations and a neighbor set of $\sim$1.17M single-edit pairs that enable fragment-level learning and controllable reasoning; and (iv) comprehensive experiments and ablations demonstrating strong results compared to baselines and clear additive gains from each component.

\section{Related Work}
\paragraph{Molecular Generation with Property Control.}
Deep generative models have been widely applied to molecular design, leveraging graph or sequence-based representations such as SMILES. Early works include VAEs \citep{gomez2018vae} and GANs such as MolGAN \citep{decao2018molgan}, followed by graph-based models like GCPN \citep{you2018gcpn}, GraphAF \citep{shi2020graphaf}, and MoFlow \citep{zang2020moflow}. Spanning Tree Graph Generation (STGG)~\citep{ahn2021spanning,jolicoeurmartineau2025anypropertyconditionalmoleculegenerationselfcriticism} shows promising performance in multi-objective optimization by combining sequential modeling with structured tree-based molecule representation. Reinforcement learning approaches, e.g., MolDQN \citep{zhou2019moldqn} and GFlowNets~\citep{bengio2023gflownet,jain2023multi}, enable property-driven optimization, often with multi-objective extensions for QED, LogP, and SA. However, these single-agent methods struggle to exactly satisfy multiple numeric constraints, reflecting exploration–exploitation trade-offs.


\paragraph{LLMs for Molecular Design and Reasoning.}
Large language models (LLMs) such as ChemGPT \citep{Frey2023NeuralSO}, ChemBERTa \citep{irwin2022ChemBERTa}, MolT5 \citep{edwards2022molt5}, ChemFM~\citep{cai2024foundation}, and Chemformer \citep{Irwin2021ChemformerAP} capture chemical syntax and semantics, enabling general-purpose molecular generation. While expressive, they remain limited in precise numerical reasoning and property control. Chain-of-Thought prompting \citep{wei2022cot} improves interpretability and multi-step reasoning in LLMs, and analogous strategies have been suggested for molecules \citep{Jin2024GraphCA,Jang2024StructuralRI,Zheng2024CriticCoTBT}, aligning with human-in-the-loop frameworks. Yet, exact satisfaction of multiple physicochemical constraints remains challenging. Recent work such as Instruction Multi-Constraint Molecular Generation \citep{zhou2025instruction} demonstrates that LLMs can satisfy multiple property constraints through teacher–student supervised training and interval-based conditioning. However, these methods primarily operate within bounded property ranges and are not based on reinforcement learning for multi-objective optimization limiting their exploration abilities.

\paragraph{Multi-Agent Planning and Reasoning in Molecule Design.}
Agent-based systems have long been studied in robotics, distributed AI, and resource allocation \citep{wooldridge2009mas,weiss1999mas}. In molecule design, however, most AI-driven approaches remain single-agent, where a single generative model is guided by property predictors. Recent work has begun to explore multi-agent systems that decompose the design process into specialized roles, such as generation, property evaluation, and refinement, by enabling cooperation or hierarchical coordination, these systems can improve exploration efficiency and controllability. For example, recent works like Prompt-to-Pill \citep{vichentijevikj2025prompt}, ROBIN \citep{ghareeb2025robin}, DrugAgent~\citep{liu2024drugagent}, Honeycomb~\citep{zhang2024honeycombflexiblellmbasedagent} and ChemCrow~\citep{m2024augmenting} have demonstrated the power of this multi-agent paradigm. Building on this line of research, we introduce a retrieval-augmented multi-agent reasoner that iteratively constructs locally optimal prototypes before refinement. This allows our system to combine in-distribution retrieval with domain knowledge to improve controllability under numeric property constraints.

\paragraph{Policy Optimization for Multi-Property Objectives.}
Reinforcement learning provides a foundation for molecular optimization. Classical policy-gradient methods such as REINFORCE \citep{williams1992reinforce} and proximal policy optimization (PPO) \citep{schulman2017ppo} have been adapted to molecule design. MolDQN \citep{zhou2019moldqn}, for example, leverages Q-learning for multi-objective optimization. However, these approaches face difficulties in balancing multiple numeric objectives precisely. 
Group Relative Policy Optimization (GRPO)~\citep{shao2024deepseekmathpushinglimitsmathematical,zhang2025critiquegrpoadvancingllmreasoning}, originally introduced for preference-based learning and RLHF, optimizes policies via group-relative advantages that reward candidates outperforming their peers. While GRPO and its modified versions are well known for strengthening LLM reasoning \citep{deepseekai2025deepseekr1incentivizingreasoningcapability}, we are the first to adapt it to numerically conditioned generation, integrating fragment-level refinement and controllable multi-hop optimization within the generation loop. This yields a principled reinforcement-learning framework for satisfying numeric multi-property targets.

\captionsetup{skip=4pt}
\begin{figure*}[t]
    \centering
    \includegraphics[width=\textwidth]{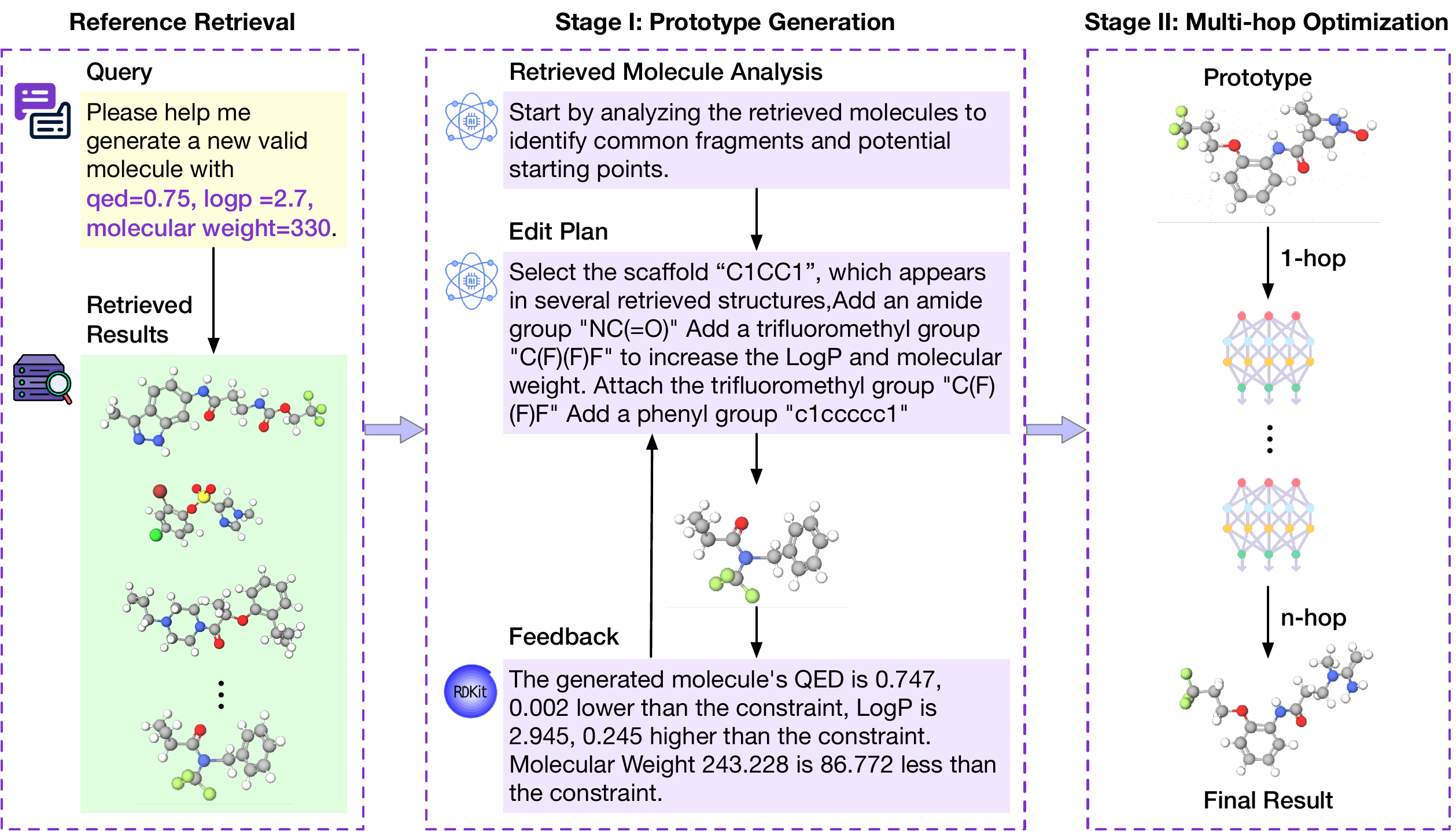}
    \caption{The flow chart of M$^{4}$olGen. The first two blocks involve \textbf{Retrieval and Prototyping}, where molecular candidates are first retrieved based on the given constraints (QED, LogP, MW) and then analyzed by a local reasoner to extract constraints, analyze retrieved molecules, and propose an editing plan based on evaluator's feedback to generate prototypes iteratively. The third block describes \textbf{Multi-Hop Optimization}, where the prototypes are optimized through one-hop and n-hop controllable editing steps by the molecule optimizer trained by GRPO.}
    \vspace{-10pt}
    \label{fig:retrieval-optimization}
\end{figure*}
\vspace{-5pt}
\section{Methodology}
\vspace{-5pt}
We propose M$^{4}$olGen (Figure~\ref{fig:retrieval-optimization}) a multi-stage, goal-conditioned framework for constrained molecular generation that casts numeric targets (QED, LogP, MW, HOMO, LUMO) as a verifiable distance-to-target objective over an actionable fragment-edit space. In the following methodology part, we will describe our system using the first set of properties (QED, LogP, MW) as an example for clearer presentation.  Stage~I performs retrieval-augmented prototyping: a local reasoner edits fragments using in-distribution exemplars and RDKit feedback to place a candidate near the feasible region. Stage~II applies a GRPO-trained fragment-level optimizer in a multi-hop manner to minimize the distance-to-target while regulating edit complexity and deviation from the starting structure. Trained on a large, property-annotated neighbor dataset, M$^{4}$olGen generalizes across target tuples and delivers precise, simultaneous control of set 1 (QED, LogP, MW) and set 2 (HOMO, LUMO).

\subsection{Stage~I: Prototype Generation with Retrieve-augmented Multi-agent Reasoning}

The objective of Stage I is to generate a chemically valid prototype \( m_{\mathrm{local}} \) that serves as a high-quality starting point for numeric optimization. This is accomplished via a collaborative multi-agent framework that decomposes the input query, retrieves similar molecules from a large database, and incrementally proposes fragment-level edits based on domain knowledge.

\paragraph{Query Interpretation.}
Given a natural-language request $q$ (e.g., \textit{“Generate a molecule with QED=$0.75$, LogP=$2.7$, MW=$310$”}), this module extracts the exact numeric targets for each property and returns a target property vector
\begin{equation}
\begin{aligned}
\mathbf{p}_{\mathrm{tgt}}
= \big(p_{\mathrm{QED}},\, p_{\mathrm{LogP}},\, p_{\mathrm{MW}}\big),
\\
p_{\mathrm{QED}}\!\in[0,1],\;
p_{\mathrm{LogP}}\!\in\mathbb{R},\;
p_{\mathrm{MW}}\!>\!0.
\end{aligned}
\end{equation}
We use \emph{$\mathbf{p}$ for “properties”} and the subscript “tgt” to denote targets. A rule-based parser identifies numeric constraints and synonyms (e.g., “molecular weight”, “MW”).

\paragraph{Reference Retrieval.}
Given the target property vector $\mathbf{p}_{\mathrm{tgt}}$, we query a large annotated molecule corpus $\Omega$ to obtain a set of \emph{reference molecules} that lie close to the targets under per–property tolerances:
\begin{equation}
\begin{aligned}
\mathcal{M}
= \big\{\, m \in \Omega \;:\; |\,p_i(m) - p_{i,\mathrm{tgt}}\,| \le \epsilon_i \;\;\\ \forall\, i \in \{\mathrm{QED},\mathrm{LogP},\mathrm{MW}\} \big\}.
\end{aligned}
\end{equation}
Here $p_i(m)$ denotes the $i$-th property of molecule $m$ (computed via RDKit), and $\epsilon_i$ are small, property-specific tolerant ranges, e.g., $\pm 0.05$ for QED (0--1 scale), $\pm 0.5$ for LogP (small medicinally meaningful shift), and $\pm 25$\,Da for MW. They are chosen to be tight enough to keep the references in-distribution yet broad enough to ensure sufficient references. The retrieved references are then used to \emph{anchor} Stage~I: they provide in-distribution exemplars that guide fragment-level edits, constrain the search toward the feasible region, and seed candidate/neighbor structures consumed by the multi-hop optimizer in Stage~II.

\paragraph{Prototype Reasoner.}
This LLM-driven module proposes stepwise, fragment-level edits to turn an initial seed (either ``start from scratch'' or molecules sampled from the reference set $\mathcal{M}$) into a high-quality \emph{prototype} close to the target. At iteration $t$, the reasoner selects an action $a_t \in \{\texttt{replace}, \texttt{add}, \texttt{remove}\}$ and applies it to obtain a new intermediate molecule along with the previous trajectory:
\begin{align}
m_t \;=\; \mathrm{Edit}\!\left(m_{t-1};\, a_t\right),
\qquad
m_t \in \mathcal{M}_{\mathrm{valid}},
\end{align}
where $\mathcal{M}_{\mathrm{valid}}$ denotes RDKit-parseable structures that pass basic valence and sanity checks. Decisions are guided by three information sources: (i) \emph{reference molecules} $\mathcal{M}$ retrieved near the target, (ii) an \emph{experience pool} of prior edits (neighbor pairs/trees) summarizing successful local transformations, and (iii) \emph{property feedback} (QED/LogP/MW) computed by RDKit on every candidate. The reasoner stops early when the distance-to-target falls below a threshold $\tau$ or when a maximum number of steps $T_{\max}$ is reached. The final result of this process is denoted as $m_{\mathrm{local}}$.

\paragraph{Validity and Error Estimation.}
Given the current prototype $m_{\mathrm{local}}$, we compute per–property deviations from the targets: 
\begin{equation}
\begin{aligned}
\Delta_{\mathrm{i}}(m) = \big|\,p_{i}(m_{\mathrm{local}}) - p_{i,\mathrm{tgt}}\,\big|,\\ i\in \{\text{QED, LogP, MW}\},
\end{aligned}
\end{equation}
and aggregate them into a distance-to-target objective $E(m)=\sum_i w_i\,\Delta_i(m)$ with property-specific weights. These errors are fed into the Stage~II optimizer prompt to enable targeted refinement.

\paragraph{Stage~I Objective.}
Formally, Stage~I seeks a valid prototype along the reasoning trajectory $\mathcal{G}=\{m_0,\dots,m_T\}$ that minimizes the distance-to-target:
\begin{equation}
\begin{aligned}
m_{\mathrm{local}} = \arg\min_{m \in \mathcal{G} \cap \mathcal{M}_{\mathrm{valid}}} & \sum_{i \in \{\text{QED, LogP, MW}\}} \\
& w_i \,\big|\,p_i(m) - p_{i,\mathrm{tgt}}\,\big|.
\end{aligned}
\end{equation}
The algorithm is stated in Appendix \ref{sec:algorithm}. This stage reliably moves the candidate into the feasible region by leveraging relevant molecules, past experience, and tool feedback. However, a multi-agent reasoner that is not further trained has a performance limitation on fine-grained, precise multi-property control. Stage~II addresses this by applying a GRPO-trained, fragment-level optimizer in a controlled multi-hop fashion to further reduce the total error $E(m)$ while regulating edit complexity and deviation from the starting structure.

\subsection{Stage~II: Fragment-Level Optimization via GRPO (Multi-Hop Extension)}
While Stage I reliably moves a candidate toward the feasible region, precise control of multiple numeric properties (e.g., QED, LogP, MW) remains difficult for text-only planning because LLMs have difficulty dealing with numeric-related design and lack a mechanism to explicitly minimize the distance to target values. Our insight is to treat refinement as an optimization problem over an actionable fragment-edit space with fast, verifiable feedback from chemistry oracles. We therefore train an optimization policy with \textbf{GRPO} (Group Relative Policy Optimization) \citep{deepseekai2025deepseekr1incentivizingreasoningcapability} because its group-wise, rank-based updates are stable and sample-efficient without ground-truth demonstrations, and because it can directly optimize a reward that faithfully encodes the numeric targets. RDKit oracles provide the property feedback at each step, making the reward precise and inexpensive to evaluate.
\vspace{-5pt}
\paragraph{Fragmentization and Action Space.}
Let $m_0 \coloneqq m_{\mathrm{local}}$ be the prototype from Stage~I. We decompose molecules into chemically meaningful building blocks using \textbf{BRICS} (Break Retrosynthetically Interesting Chemical Substructures)~\citep{degen2008art}, a rule-based scheme that cuts retrosynthetically plausible bonds formed or broken during synthetic processes, leading to fragments that are synthetically accessible and chemically meaningful. This yields fragments that support localized edits, preserve validity, and keep the search space tractable where $\Phi(m)$ is the fragment set for molecule $m$ and $f$ are the fragments:
\begin{align}
\Phi(m) \;=\; \{\,f_1,\dots,f_k\,\}.
\end{align}
At hop $h \in \{1,\dots,H\}$, the optimizer selects one fragment-level action $a_h \in \{\texttt{add},\texttt{remove},\texttt{replace}\}$ and applies it to obtain a new candidate
\begin{align}
m_h \;=\; \mathrm{Edit}(m_{h-1}; a_h),
\qquad
m_h \in \mathcal{M}_{\mathrm{valid}},
\end{align}
where $\mathcal{M}_{\mathrm{valid}}$ denotes RDKit-parseable structures that pass basic valence and sanity checks. A hop budget $H$ controls structural complexity and deviation from the starting structure.

\paragraph{Optimizer and Input Representation.}
Our optimizer $\mathcal{O}_{\phi}$ is a sequence model (an LLM policy) fine-tuned with GRPO on our neighbor-pair corpus of single-fragment edits. Following \cite{guevorguian2024smallmoleculeoptimizationlarge}, we extend the tokenizer with \texttt{<SMILES>}, \texttt{</SMILES>}, \texttt{<QED>}, \texttt{</QED>}, \texttt{<LogP>}, \texttt{</LogP>}, \texttt{<MW>}, \texttt{</MW>} so that molecules and targets are explicit in the prompt. At each hop, the policy conditions on $(m_{h-1}, \Phi(m_{h-1}), \mathbf{p}_{\mathrm{tgt}})$ and proposes one edited molecule; after $H$ hops we return $m^* \coloneqq m_H$.

\paragraph{Reward and GRPO Objective.}
We define a distance-to-target objective and convert it to a scalar reward using fast RDKit oracles:
\begin{equation}
\begin{aligned}
E(m) \;=\; \sum_{i \in \{\mathrm{QED},\mathrm{LogP},\mathrm{MW}\}} w_i\,\big|\,p_i(m) - p_{i,\mathrm{tgt}}\,\big|
,
\\
r_{\mathrm{prop}}(m) \;=\; 1 - E(m).
\end{aligned}
\end{equation}
The full reward combines format, property, diversity, and validity terms:
\begin{equation}
\begin{aligned}
R(m) \;=\; 
\underbrace{r_{\mathrm{format}}(m)}_{\text{valid SMILES / instruction}}
\;+\;
\underbrace{r_{\mathrm{prop}}(m)}_{\text{scaled property match}}\\
\;-\;
\underbrace{r_{\mathrm{repeat}}(m)}_{\text{repetition penalty}}
\;-\;
\underbrace{r_{\mathrm{invalid}}(m)}_{\text{RDKit parse / valence penalty}}.
\end{aligned}
\end{equation}
Here $w_i$ are \emph{weights} that balance units and priorities. We also optionally regularizes complexity (e.g., hop count or similarity). GRPO samples a group of candidates, ranks them by $R(m)$, get the normalized rewards from reward functions, and updates the policy to increase the likelihood of higher-ranked edits while discouraging weaker ones. This group-relative signal is robust under noisy rewards and directly steers the policy toward exact numeric targets.
\vspace{-5pt}
\paragraph{Multi-hop Refinement and Control.}
Applying the optimizer in a controlled multi-hop manner enables gradual, interpretable refinement: small, local edits accumulate to tighten requirement satisfaction, while the hop budget and regularizers bound complexity and deviation from the prototype. In practice, a modest $H$ suffices to reliably reduce $E(m)$ thanks to fragment locality and fast RDKit evaluation, and the same mechanism supports adaptive planning and curriculum-style difficulty scaling during training and evaluation.

\subsection{Automated Synthesis of Reasoning Dataset}
\vspace{-5pt}
To train an optimizer that not only generates strings, but reasons about edits, we require a corpus that (i) couples each molecule with reliable physicochemical properties, (ii) exposes an actionable fragment space (fragments and how they connect), and (iii) provides neighbor relations so we can supervise single-step edits and assemble multi-hop reasoning chains. This enables reward-driven refinement under exact numeric targets. More details are shown in Appendix \ref{sec:dataset}

\vspace{-5pt}
\section{Experiment}
\vspace{-5pt}
We conduct the experiments for the following three claims. \textbf{(C1) Precise multi-property control:} we benchmark M$^{4}$olGen against strong LLMs and graph methods under identical compute budgets, reporting per-property MAE and a normalized total error to demonstrate simultaneous control of QED/LogP/MW and HOMO/LUMO. \textbf{(C2) Necessity and effectiveness of the two-stage design:} we perform ablations that toggle retrieval in Stage~I and vary the GRPO optimizer hops (1/2/3), to show that retrieval-augmented prototyping plus multi-hop refinement is required for tight numeric alignment. \textbf{(C3) Generalization without per-target retraining \& controllable edit complexity:} we uniformly sample 100 target tuples across admissible ranges, run 10 trials per tuple/baseline (best-of-10 under a fixed budget), and analyze performance as a function of hop budget, establishing broad generalization and explicit control of deviation from the prototype.

\subsection{Experimental Setup}
\paragraph{Training Details.}
In Stage I, we employ GPT-4o~\citep{openai_gpt4o_systemcard} or 
Qwen3-14B-chem-dyn-tokenizer~\citep{qwen3_14b_chem_dyn_tokenizer} as the prototype-reasoning LLM. For the Stage II training,  we select ChemDFM-v1.5-8B~\citep{Zhao_2025} as the base model, which achieves overall great performance among chemical generation tasks among models within 8B. We first train ChemDFM-v1.5-8B for 5000 steps with supervised fine-tuning to strength the output format as cold start. This can accelerate the convergence speed for the following GRPO training since the reward function can get effective feedback sooner than randomly exploring the format first. Then the model is trained for 37,500 steps with GRPO. The scalars we choose to normalize errors for the reward function are \(\alpha_q\)=1, \(\alpha_l\)=6, \(\alpha_w\)=100, as we consider error values 1 in QED, 6 in LogP and 100 in MW as the maximum thresholds. There is no scalar for HOMO and LUMO. We directly use the MAE in the reward function.  These scalars are flexible to tune depending on personal usage. The invalidity penalty and wrong format penalty are both -10 while the repetition penalty is accumulated by 0.1 for each time. At each step, we sample \(8\) candidates using stochastic decoding (temperature \(T=1.0\), top-\(p=0.9\), top-\(k=50\)). The model was trained to convergence on a single NVIDIA A100 (40\,GB).
\vspace{-5pt}
\paragraph{Baselines.}
We aim to investigate the power of LLMs for generating new molecules under precise constraints. Thus, most of the baselines we choose are LLMs. In the LLM-based solutions, we have gpt-4.1~\citep{openai_gpt41_2025_docs}, Gemini-Flash~\citep{google_gemini_15_flash}, claude-haiku~\citep{anthropic_claude3_haiku_2024}, gpt-4o-2024-05-13(latest version)~\citep{openai_gpt4o_20240513}, SmileyLlama-8B~\citep{cavanagh2025smileyllamamodifyinglargelanguage} and DrugAssist-7B~\citep{ye2023drugassistlargelanguagemodel}. They cover most commonly used commercial models and generation-oriented fine-tuned chemical LLMs including the SFT (Supervised Fine-Tuning) and DPO (Direct Preference Optimization) techniques.
In addition to LLM baselines, we also compare to commonly used graph-based and hybrid algorithms. STGG+~\citep{jolicoeurmartineau2025anypropertyconditionalmoleculegenerationselfcriticism}, which is a strong autoregressive generative model that uses spanning tree-based graph generation to perform multi-property conditional generation. We also include a graph genetic algorithm (Graph GA)~\citep{jensen2019graph}, which is based on target-specific optimization; for each target tuple we run it from scratch with oracle calls of 500 and 1000 (denoted GA-500 and GA-1000). There is no LLM-based solution achieving decent performance (below 3eV as total MAE) on HOMO-LUMO constrained generation. Thus, we only include Graph GA in the HOMO-LUMO baselines as it is the strongest solution for reward-aware scenario.
\vspace{-5pt}
\paragraph{Metrics.}
We compute all the properties of generated samples and compare them with the target to get the MAE (mean absolute error). MAE is commonly used among molecular generation and design benchmarks~\citep{wu2018moleculenet}. However, for the multi-objective optimization task that we address, it is necessary to have a normalized total error so that we can directly determine which candidate is better. Different properties have different ranges, and individual properties need to be normalized to the same range for multi-objective molecule design \citep{LUUKKONEN2023102537}. Therefore, we normalize the error by dividing QED error by 1, LogP error by 10 and MW error by 700 since QED range is from 0 to 1, LogP range is from -10 to 10 and most in-distribution MW range is from 100 to 800. Note that the normalizer for each error can be tuned when dealing with custom distribution or specific-property-preferred generation. Besides the whole range normalization, we also add the scalars we used for the optimizer's GRPO training (1 for QED, 6 for LogP and 100 for MW). No scalars are needed in HOMO-LUMO experiments as their value ranges are the same. Beyond MAE, we assess \textbf{set quality}. \emph{Uniqueness} is the fraction of distinct molecules among the outputs (measured via canonical SMILES), indicating the absence of duplicates. \emph{Diversity} measures how dissimilar the set is on average, computed from ECFP4 fingerprints~\citep{rogers2010extended} with the Tanimoto similarity (higher diversity means broader exploration of chemical space).

\begin{table*}[ht]
\centering
\caption{Overall metrics across methods. Lower is better for error metrics; higher is better for diversity and uniqueness. Best per column in \textbf{bold}; second best underlined.}
\label{tab:structured-errors}
\resizebox{\textwidth}{!}{
\begin{tabular}{
    l 
    S[table-format=1.3] 
    S[table-format=1.3] 
    S[table-format=3.3] 
    S[table-format=1.3] 
    S[table-format=1.3] 
    S[table-format=1.3] 
    S[table-format=1.2]
    }
\toprule
\textbf{{Method}} & \textbf{{QED err}} & \textbf{{logP err}} & \textbf{{MW err}} & \textbf{{Scaled total err}} & \textbf{{Norm. total err}} & \textbf{{Diversity}} & \textbf{{Uniqueness}}\\
\midrule
\multicolumn{5}{l}{\textbf{LLMs}}\\
\addlinespace[2pt]
gpt-4.1    & 0.115 & 0.697 & 49.182 &0.723 & 0.255 & 0.823& 1.0\\
gpt-4o-2024-05-13  & 0.115 & 0.847 &  60.203 &0.858 & 0.285 & 0.868 & 1.0\\
Gemini-2.5-Flash       & {\bfseries 0.078} & 0.974 & 86.174 &1.102 & 0.299 & 0.842 & 0.97\\
Claude-3.7-Sonnet  & 0.104 & 1.025 & 39.583 &0.671 & 0.263 & 0.868 &1.0\\
Claude-3.5-haiku       & 0.117 & 1.174 & 46.904 &0.782& 0.301 & 0.791 & 1.0\\
SmileyLlama-8B        & 0.374 & 2.385 & 196.235 &2.734 & 0.893 &0.853 &1.0 \\
DrugAssist-7B      &0.176	&2.44	&165.047	&2.233	&0.656&	0.845	&0.38\\
\midrule
\addlinespace[4pt]
\multicolumn{5}{l}{\textbf{Graph algorithms}}\\
\addlinespace[2pt]
STGG+     & \underline{0.079} & 0.418 & 23.56 & 0.385 &\underline{0.155} &0.879&1.0 \\
Graph GA-500 &0.131	&0.806	&15.016	&0.415	&0.233&	\underline{0.884}&	1.0 \\
Graph GA-1000 &0.123	&0.529	&{\bfseries 7.95}	&0.291	&0.187&	{\bfseries 0.886}&	1.0\\
\midrule
\addlinespace[4pt]
\multicolumn{5}{l}{\textbf{Our methods}}\\
\addlinespace[2pt]
1-hop-GPT4o              & 0.130 & 0.423 & 10.404 & 0.305& 0.187 &0.879 & 1.0 \\
2-hop-GPT4o              & 0.111 & 0.339 & 10.489 &0.272 & 0.160 &0.883 & 1.0\\
3-hop-GPT4o              & 0.103 & \underline{0.284} & \underline{9.799} &{\bfseries 0.249} & {\bfseries 0.146} & \underline{0.884} & 1.0\\
1-hop-Qwen              & 0.152 & 0.319 & 16.176 & 0.367& 0.204 &0.869 & 1.0 \\
2-hop-Qwen              & 0.132 & {\bfseries 0.209} & 11.864 & 0.285& 0.168 &0.876 & 1.0 \\
3-hop-Qwen              & 0.120 & 0.237 & 10.365 & \underline{0.263} & 0.159 &0.878 & 1.0 \\
\bottomrule
\end{tabular}
\vspace{-10pt}
}
\end{table*}

\begin{table*}[t]
\centering
\small
\caption{Performance comparison across methods on HOMO/LUMO–constrained molecular generation.}
\label{tab:homolumo_results}
\begin{tabular}{lccccc}
\toprule
\textbf{Method} 
& \textbf{HOMO Error} 
& \textbf{LUMO Error} 
& \textbf{Total Error} 
& \textbf{Diversity} 
& \textbf{Uniqueness} \\
\midrule
Graph GA-500        & 0.369 & 0.396 & 0.765 & 0.881 & 0.957 \\
Graph GA-1000        & 0.251 & 0.353 & 0.604 & 0.894 & 1.0 \\
1-hop    & 0.301 & 0.239 & 0.540 & 0.874 & 1.0 \\
2-hop    & 0.112 & 0.115 & 0.227 & 0.867 & 1.0 \\
3-hop    & \textbf{0.060} & \textbf{0.095} & \textbf{0.155} & 0.862 & 1.0 \\
\bottomrule
\end{tabular}
\end{table*}
\vspace{-3pt}
\subsection{Results and Analysis}
\vspace{-2pt}
\paragraph{Protocol.} GRPO is ground-truth–free and reward-based, so performance is not tied to a particular training distribution. To test generalization, we uniformly sample 100 target tuples $(\mathrm{QED},\mathrm{LogP},\mathrm{MW})$ and $\mathrm{HOMO}, \mathrm{LUMO}$ across admissible ranges. For each tuple and each baseline we run 10 independent trials under the same compute budget and report the \emph{best-of-10}. Across settings, our normalized total error (NTE) decreases monotonically with hop count shown in both Table~\ref{tab:structured-errors} and Table~\ref{tab:homolumo_results}.

\paragraph{Main Results.} Table~\ref{tab:structured-errors} compares LLMs, graph baselines, and our method. Our best configuration (3-hop-GPT-4o) attains the lowest \textbf{NTE} (normalized total error) of \textbf{0.146}, improving over the strongest commercial model (GPT-4.1, $0.255$) by \textbf{42.7\%} and outperform the best non-LLM baseline (STGG+). Meanwhile, Qwen-based configuration also achieves competitive performance with second best NTE($0.159$). This points out that our method does not rely on commerical models. Per metric, we obtain the best \textbf{logP} error (\textbf{0.209}) and the second-best \textbf{MW} error (\underline{9.799}; GA-1000 is \textbf{7.95}). Relative to STGG-50$\times$, our 3-hop reduces logP from $0.566$ to $0.284$ (\textbf{49.8\%}) and MW from $63.917$ to $9.799$ (\textbf{84.7\%}); STGG-50$\times$ achieves the best QED ($\mathbf{0.050}$), while ours remains competitive ($0.103$). Diversity and uniqueness are high (Div $\approx 0.884$, Uniq $=1.0$), on par with the best graph baseline (Graph GA-1000, Div $=0.886$).

Table~\ref{tab:homolumo_results} shows the performance of the generation under HOMO and LUMO constraints (unit is eV). M$^{4}$olGen significantly outperforms the Graph GA baseline. The 1-hop setting already reduces the total error to 0.540, demonstrating that a single controlled fragment edit guided by property feedback can effectively move molecules toward the target electronic profile. The 2-hop strategy reduces the total error to 0.227, achieving more than a \textbf{2×} improvement over Graph GA-1000. The 3-hop configuration achieves the best overall performance, with a total error of \textbf{0.155} and particularly low HOMO and LUMO errors (0.060 and 0.095, respectively). Notably, both HOMO and LUMO errors are simultaneously reduced, suggesting balanced multi-property optimization rather than overfitting to a single objective. While Graph-GA becomes increasingly time-consuming as oracle call increases, M$^{4}$olGen achieves comparable or better performance with nearly 90\% less inference time. Because the optimization cost is amortized after a single training phase, M$^{4}$olGen is substantially more efficient for repeated or large-scale molecule generation tasks.

\begin{table*}[!h]
\centering
\small
\caption{Ablation study on retrieval and fragment-level optimizer (lower is better).}
\label{tab:ablation}
\begin{tabular}{l S[table-format=1.3] S[table-format=1.3] S[table-format=3.3] S[table-format=1.3]}
\toprule
{\textbf{Method}} & \textbf{{QED err}} & \textbf{{logP err}} & \textbf{{MW err}} & \textbf{{Norm. total err}} \\
\midrule
Stage1 (no retrieval)           & 0.111 & 0.970 & 68.555 & 0.307 \\
Stage1 + retrieval              & \textbf{0.098} & 0.769 & 63.240 & 0.265 \\
Stage1 + retrieval + 1-hop      & 0.130 & 0.423 & \underline{10.404} & 0.187 \\
Stage1 + retrieval + 2-hop      & 0.111 & \underline{0.339} & 10.489 & \underline{0.160} \\
Stage1 + retrieval + 3-hop      & \underline{0.103} & \textbf{0.284} & \textbf{9.799} & \textbf{0.146} \\
\bottomrule
\end{tabular}
\end{table*}

\vspace{-5pt}

\subsection{Ablation Study}
\label{sec:ablation}

\begin{figure}[t]
  \centering
  \includegraphics[width=\linewidth]{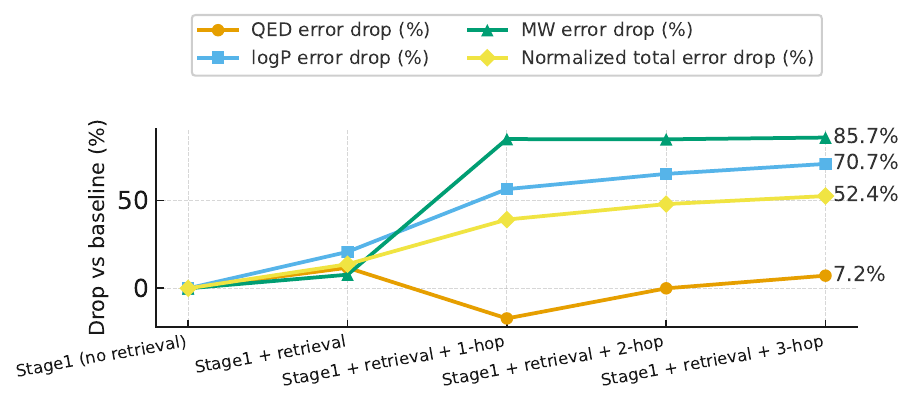}
  \caption{Ablation curves showing the \emph{drop percentage} (higher is better) of each error metric relative to the no-retrieval baseline across methods. Curves are shown for QED, logP, MW, and the normalized total error.}
  \label{fig:ablation-drop}
\end{figure}

\paragraph{Interpretation.} We ablate three design choices on a held-out set: (i) Stage~I without retrieval (baseline), (ii) Stage~1 with retrieval, and (iii) Stage~I with retrieval followed by a fragment-level optimizer using 1/2/3 hops. We report per-property errors (QED, logP, MW) and the normalized total error (\(e_{\mathrm{norm}}=\lvert \Delta \mathrm{QED}\rvert + \lvert \Delta \log P\rvert /10 + \lvert \Delta \mathrm{MW}\rvert /700\)) in Table~\ref{tab:ablation}. For visualization, we plot the \emph{drop percentage} relative to the no-retrieval baseline,
\[
\text{drop}(m) = \frac{e_{\text{base}} - e_m}{e_{\text{base}}}\times 100\%,
\]
for each metric and method (Figure~\ref{fig:ablation-drop}).

\paragraph{Effect of Retrieval.} Adding retrieval already yields consistent gains: the normalized total error drops by \textbf{13.7\%} (\(0.307 \to 0.265\)), driven primarily by improvements in logP (\(\mathbf{20.7\%}\) drop) and MW (\(7.8\%\) drop). Retrieval also gives the best stand-alone QED error among non-optimized variants (\(\mathbf{0.098}\), \(\mathbf{11.7\%}\) drop).

\paragraph{Effect of the Fragment-level Optimizer.} Introducing the optimizer produces the largest improvements, especially on MW. Moving from retrieval-only to 1/2/3 hops reduces MW error from \(\sim 63\) to \(\sim 10\) (\(\mathbf{84.9\%}\) drop vs. baseline), and steadily improves logP (drops of \(56.4\%\), \(65.1\%\), and \(\mathbf{70.7\%}\)). The overall normalized error decreases monotonically with more hops: \(0.187\) (1-hop, \(39.1\%\) drop), \(0.160\) (2-hop, \(47.9\%\) drop), and \(\mathbf{0.146}\) (3-hop, \(\mathbf{52.4\%}\) drop). QED exhibits a small regression at 1-hop (as expected when trading off multi-objective targets), but recovers by 3-hop to a \(7.0\%\) drop versus baseline.

\paragraph{Takeaway.} Retrieval is a strong enabler, and the fragment-level optimizer is essential for precise multi-property alignment, culminating in the best overall performance with the 3-hop setting.

\vspace{-3pt}
\section{Conclusion}
\vspace{-2pt}
We introduced \textbf{M$^{4}$olGen}, a two-stage, fragment-level framework for \emph{precise, property-constrained} molecular generation and a large, reasoning-ready dataset (BRICS fragments with neighbor pairs and measured property deltas) for training. Across QED, $LogP$, and MW and HOMO, LUMO targets, M$^{4}$olGen attains the lowest normalized total error among LLM and graph baselines, with monotonic gains as hop count increases, and maintains validity, uniqueness, and diversity. Taken together, these results validate our design choices and demonstrate the method’s potential to scale to richer objectives.

\section*{Limitations}

While promising, our study is limited by its reliance on computed properties (e.g., RDKit estimators) and by the narrow property set evaluated (QED, LogP, MW, HOMO, LUMO). This work serves as an initial validation of applying GRPO to explore discrete chemical action spaces for precise property control, where rapid reward feedback and fundamental physicochemical objectives are essential for stable training. Also, although deeper hops continue to improve performance, they incur substantially higher computation cost with diminishing returns and this is a practical design trade-off.

\section*{Acknowledgment}
The experiments were enabled in part by computational resources provided by Digital Research Alliance of Canada and Lambda Cloud. We acknowledge the support from the Samsung SAIT-Mila Collaboration Grant and the Canada CIFAR AI Chair program.

\bibliography{custom}

\appendix

\newpage
\section{Appendix}
\label{sec:appendix}
\subsection{Dataset Details}
\label{sec:dataset}
We combine all the molecules from ZINC~\citep{irwin2005zinc}, CHEMBL~\citep{gaulton2012chembl} and MOSES \citep{polykovskiy2020molecularsetsmosesbenchmarking} together, filter and delete the duplicates. From each molecule we obtain its SMILES, molecular formula, QED, logP, and molecular weight computed with RDKit and HOMO, LUMO with our pretrained evaluator based on DimeNet++~\citep{gasteiger2022fastuncertaintyawaredirectionalmessage}. We further derive a fragment decomposition and an inter-fragment connectivity map (identifying the bonds between fragments). The final dataset contains 2,945,596 molecules and, to the best of our knowledge, is the largest resource coupling molecular properties with fragment-based structural annotations. 

\begin{figure*}[t]
    \centering
    \includegraphics[width=0.95\textwidth]{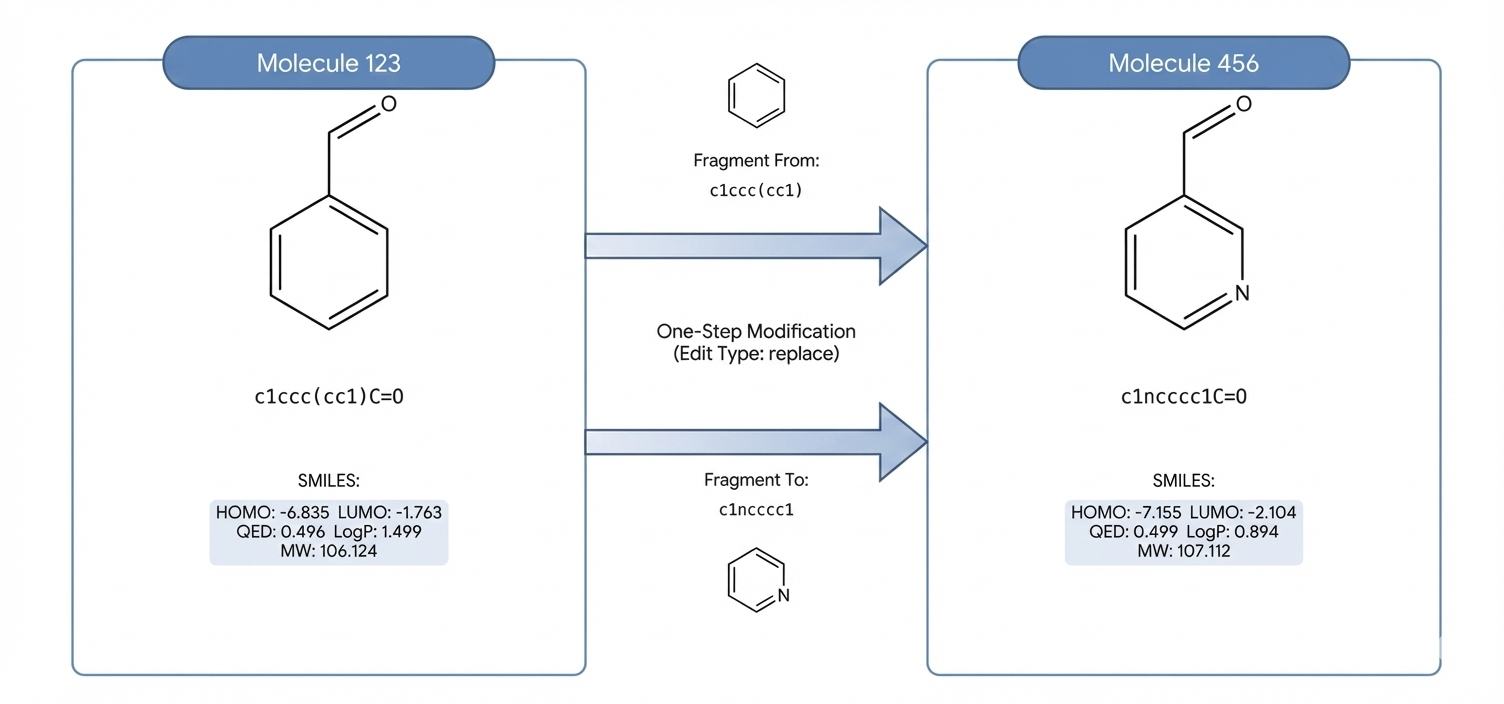}
    \caption{The demo of nodes and edges of molecule neighbor reasoning dataset.}
    \label{fig:dataset}
    \vspace{-10pt}
\end{figure*}

Starting from our unified corpus, we build a reasoning-ready resource through an automated pipeline: (i) \textbf{standardize \& deduplicate} molecules via RDKit canonical SMILES, neutralize, and enforce valence/aromaticity sanity checks; (ii) \textbf{annotate properties} (QED, LogP, MW) with RDKit and (HOMO, LUMO) with pretrained DimeNet++; (iii) \textbf{fragmentize} each molecule with BRICS to obtain a fragment multiset $\Phi(m)$ and an inter-fragment connectivity map (which fragments are joined and at which bonds), yielding an actionable edit space; (iv) \textbf{construct neighbor pairs} by scanning for molecules that differ by exactly one fragment-level edit (\texttt{add}/\texttt{remove}/\texttt{replace}), while enforcing edit sanity (e.g., element-count conservation for \texttt{replace}) and RDKit validity for the edited product; and (v) \textbf{label supervision} by recording the edit type, edited fragments, and signed property deltas $(\Delta\mathrm{QED}, \Delta\mathrm{LogP}, \Delta\mathrm{MW})$, plus the distance-to-target objective used by our optimizer. This process yields a \textbf{neighbor-pair corpus of $\sim$1{,}171{,}193 single-edit pairs}. For each molecule we also materialize its \textbf{1-hop neighbor list} based on the fragment multiset edit distance as shown in figure \ref{fig:dataset}, from which we grow neighbor trees/forests. These structures serve two roles: they seed \emph{retrieval-anchored prototyping} in Stage~I and provide \emph{experience-based, reward-compatible supervision} for GRPO in Stage~II, enabling controllable multi-hop refinement under exact numeric targets. Each entry is formatted as a natural language prompt with a one-step edit answer, e.g.:

\begin{quote}
\small
Given is an intermediate molecule SMILES \texttt{<SMILES>\seqsplit{O=C(NCc1nccc2ccccc12)c1ccc[nH]c1=O}</SMILES>}, which is composed of fragments [\texttt{'C()=O', 'N', 'C', 'c1nccc2ccccc12', 'c1ccc[nH]c1=O'}]. Propose a single replace, add or remove step on fragment level that makes the new molecule's QED \texttt{<QED>0.146</QED>} lower, LogP \texttt{<LogP>0.366</LogP>} higher, and Molecular Weight \texttt{<MW>53.068</MW>} lower.
\end{quote}
\begin{quote}
\small
Replace \texttt{c1ccc[nH]c1=O} with \texttt{c1nc2nc(C)cc(C)n2n1} to form \texttt{<SMILES>\seqsplit{Cc1cc(C)n2nc(C(=O)NCc3nccc4ccccc34)nc2n1}</SMILES>.}
\end{quote}
GRPO itself does not need any ground truth for editing, but all property changes are still derived from real data to preserve distribution realism.

\subsection{Stage~I Algorithm}
\label{sec:algorithm}
\begin{algorithm}[H]
\caption{Stage~I: Local Optimal Candidate Generation via Multi-Agent Planning}
\begin{algorithmic}[1]
\Require User query \( q \), molecule database \( \mathcal{M} \), thresholds \( \epsilon_i \), max iterations \( T \)
\State \( \mathbf{P}^* \gets \text{Decomposer}(q) \)
\State \( \mathcal{N} \gets \text{Retriever}(\mathbf{P}^*, \mathcal{M}, \epsilon_i) \)
\State \( m_0 \gets \text{InitialGeneration}(q, \mathcal{N}, \mathbf{P}^*) \)
\State Initialize reasoning history \( \mathcal{H} \gets [\,] \)
\For{\( t = 1 \) to \( T \)}
    \State \( a_t \gets \text{Reasoner}(q, \mathcal{N}, \mathcal{H}, \mathbf{P}^*) \)
    \State \( m_t \gets \text{Edit}(m_{t-1}, a_t) \)
    \State \( \mathcal{H} \gets \mathcal{H} \cup \{a_t\} \)
    \If{\( \text{is\_valid}(m_t, \mathbf{P}^*, \epsilon_i) \)}
        \State \Return \( m_t \)
    \EndIf
\EndFor
\State \Return valid \( m_T \) if any
\end{algorithmic}
\end{algorithm}


\subsection{End-to-End Demo: From Local Reasoner to GRPO Refinement}
\label{sec:demo}
\paragraph{Target.}
We aim for \textbf{QED} \(\approx 0.70\), \textbf{LogP} \(\approx 1.50\), and \textbf{MW} \(\approx 300\).

\paragraph{Stage 1 — Iterative construction (LLM planner).}
The planner begins from scratch and proposes \emph{fragment-level} edits while reading back numeric feedback at each step.

\emph{Step 1.} It proposes \texttt{\seqsplit{CCN(CC)C(=O)C(C1CC1)S(=O)=O}} based on relevant molecules, reasoning that a compact sulfonamide with small rings could balance QED and LogP. Feedback shows QED \(=0.674\) (below by \(0.026\)), LogP \(=0.245\) (below by \(1.255\)), MW \(=219.306\) (below by \(80.694\)). The model decides to raise both LogP and MW.

\emph{Step 2.} To add hydrophobic mass, it benzylates the amide nitrogen, yielding \texttt{\seqsplit{CCN(Cc1ccccc1)C(=O)C(C1CC1)S(=O)=O}}. Feedback: QED \(=0.803\) (above by \(0.103\)), LogP \(=1.425\) (just \(0.075\) low), MW \(=281.377\) (still \(18.623\) low). The ring helped; MW needs a modest push upward.

\emph{Step 3.} It enlarges the small ring to a cyclohexyl to push MW/LogP: \texttt{\seqsplit{CCN(Cc1ccccc1)C(=O)C(C1CCCCC1)S(=O)=O}}. Feedback: QED \(=0.819\) (high by \(0.119\)), LogP \(=2.595\) (high by \(1.095\)), MW \(=323.458\) (high by \(23.458\)). Overshot both LogP and MW.

\emph{Step 4.} It trims to cyclopentyl: \texttt{\seqsplit{CCN(Cc1ccccc1)C(=O)C(C1CCCC1)S(=O)=O}}. Feedback: QED \(=0.820\) (high by \(0.120\)), LogP \(=2.205\) (high by \(0.705\)), MW \(=309.431\) (high by \(9.431\)). Still too heavy and too lipophilic.

\emph{Step 5.} To temper LogP/MW while retaining aromaticity, it swaps phenyl \(\rightarrow\) pyridine: \texttt{\seqsplit{CCN(Cc1ncccc1)C(=O)C(C1CCCC1)S(=O)=O}}. Feedback: QED \(=0.811\) (high by \(0.111\)), LogP \(=1.600\) (high by \(0.100\)), MW \(=310.419\) (high by \(10.419\)). Closer on LogP, MW still a bit high.

\emph{Step 6 (seed for Stage 2).} It reduces the ring to a butyl chain to lower MW/LogP: \texttt{\seqsplit{CCN(Cc1ncccc1)C(=O)C(CCC)S(=O)=O}}. Feedback: QED \(=0.764\) (high by \(0.064\)), LogP \(=1.210\) (low by \(0.290\)), MW \(=284.381\) (low by \(15.619\)). This is the best Stage-1 candidate (normalized total error \(=0.116\)) and becomes the seed for Stage 2.

\paragraph{Stage 2 — GRPO Refinement (Accepted Path with Reasoning).}
We now switch to the optimizer trained with GRPO. At each hop, we ask for a single fragment edit that moves QED/LogP/MW by specified deltas in the right directions, then accept only moves that improve the objective.

\emph{Hop 1.} From the seed \texttt{\seqsplit{CCN(Cc1ncccc1)C(=O)C(CCC)S(=O)=O}}, we request: decrease QED by \(0.064\), increase LogP by \(0.290\), and increase MW by \(15.619\).  
\textbf{Reasoning.} The model replaces the sulfone side chain with a bicyclic, more drug-like fragment to add hydrophobic mass while modulating polarity.  
\textbf{Edit.} Replace \texttt{C(=O)C(CCC)[SH](=O)=O} \(\rightarrow\) \texttt{C1=CNC(N)C(O)C=C(C)CC1=C}, producing \texttt{\seqsplit{CCN(Cc1ncccc1)C1=CNC(N)C(O)C=C(C)CC1=C}}. The move improves the objective and is \emph{accepted}.

\emph{Hop 2.} From that intermediate, we request: further decrease QED by \(0.040\), decrease LogP by \(0.386\), and decrease MW by \(14.433\).  
\textbf{Reasoning.} The optimizer softens hydrophobicity and trims mass while preserving the newly introduced scaffold connectivity.  
\textbf{Edit.} Replace \texttt{N()C1=CNC(N)C(O)C=C(C)CC1=C} \(\rightarrow\) \texttt{NC1=CNNC=CC(O)CC(C)C1}, yielding \texttt{\seqsplit{CCNC1=CNNC=CC(O)CC(C)C1Cc1ncccc1}}. This further reduces the objective and is \emph{accepted}.

\paragraph{Final Outcome.}
The best molecule along this path is \texttt{\seqsplit{CCNC1=CNNC=CC(O)CC(C)C1Cc1ncccc1}} with QED \(=0.681\), LogP \(=1.700\), MW \(=302.422\), and a normalized total error of \(0.042\). In summary, Stage 1 quickly assembled a plausible prototype with sensible fragment choices, and Stage 2 applied two targeted, GRPO-guided edits that traded off hydrophobic mass and polarity to tighten alignment with all three numeric targets.

\subsection{Use of LLMs}
Large Language Models (LLMs) were used solely for writing refinement such as grammar and syntax improvements.

\end{document}